\documentclass[letterpaper]{article} % DO NOT CHANGE THIS
\usepackage{aaai25}  % DO NOT CHANGE THIS
\usepackage{times}  % DO NOT CHANGE THIS
\usepackage{helvet}  % DO NOT CHANGE THIS
\usepackage{courier}  % DO NOT CHANGE THIS
\usepackage[hyphens]{url}  % DO NOT CHANGE THIS
\usepackage{graphicx} % DO NOT CHANGE THIS
\usepackage{natbib}  % DO NOT CHANGE THIS AND DO NOT ADD ANY OPTIONS TO IT
\usepackage{caption} % DO NOT CHANGE THIS AND DO NOT ADD ANY OPTIONS TO IT
\usepackage{algorithm}
\usepackage{algorithmic}
\usepackage{color}
\usepackage{multirow}
\usepackage{amsmath}
\usepackage{url}

\usepackage{newfloat}
\usepackage{listings}
\DeclareCaptionStyle{ruled}{labelfont=normalfont,labelsep=colon,strut=off} % DO NOT CHANGE THIS
\floatstyle{ruled}
\newfloat{listing}{tb}{lst}{}
\floatname{listing}{Listing}

\nocopyright
\title{AlignedKV: Reducing Memory Access of KV-Cache with Precision-Aligned Quantization}
\author{
    Yifan Tan\textsuperscript{\rm 1,\rm 2},
    Haoze Wang\textsuperscript{\rm 1,\rm 2},
    Chao Yan\textsuperscript{\rm 2},
    Yangdong Deng\textsuperscript{\rm 1,}\thanks{corresponding author}
}
\affiliations{
    \textsuperscript{\rm 1}School of Software, Tsinghua University\\
    \textsuperscript{\rm 2}Sunlune\\
    tanyf24@mails.tsinghua.edu.cn, wanghz\_thu18@163.com, yanchao@mail.ioa.ac.cn, dengyd@tsinghua.edu.cn
}

\usepackage{bibentry}
\begin{document}

\maketitle

\begin{abstract}
Model quantization has become a crucial technique to address the issues of large memory consumption and long inference times associated with LLMs. Mixed-precision quantization, which distinguishes between important and unimportant parameters, stands out among numerous quantization schemes as it achieves a balance between precision and compression rate. However, existing approaches can only identify important parameters through qualitative analysis and manual experiments without quantitatively analyzing how their importance is determined. We propose a new criterion, so-called “precision alignment”, to build a quantitative framework to holistically evaluate the importance of parameters in mixed-precision quantization. Our observations on floating point addition under various real-world scenarios suggest that two addends should have identical precision, otherwise the information in the higher-precision number will be wasted. Such an observation offers an essential principle to determine the precision of each parameter in matrix multiplication operation. As the first step towards applying the above discovery to large model inference, we develop a dynamic KV-Cache quantization technique to effectively reduce memory access latency. Different from existing quantization approaches that focus on memory saving, this work directly aims to accelerate LLM inference through quantifying floating numbers. The proposed technique attains a 25\% saving of memory access and delivers up to 1.3× speedup in the computation of attention in the decoding phase of LLM, with almost no loss of precision.
\end{abstract}

\begin{links}
    \link{Code}{https://github.com/AlignedQuant/AlignedKV}
\end{links}

\section{Introduction}

Large Language Models (LLMs) are increasingly becoming a fundamental force, revolutionizing human life.The emergent capability of LLMs, however, depends on their huge number of parameters, which have to incur significant overhead in terms of large memory consumption, intensive computational cost, and high memory latency. As an effective way to deal with the abovementioned problems, model quantization has attracted heavy research effort \citep{zhu2023survey, yuan2024llm}. In these works, some methods, e.g., integer-only networks, aim to improve computational speed, while others are dedicated to reducing memory latency by cutting down the volume of memory access. Our approach belongs to the latter category.

An essential problem for quantization is to achieve a high compression ratio and at the same time maintain a sufficient level of precision. Among various quantization methods, mixed-precision quantization stands out for its ability to reach such a balance. Researchers \citep{dettmers2022gpt3,lee2024owq} already notice that not all values in a model are equally important, and quantizing important values into higher precision can decrease the precision loss caused by quantization. However, there are three questions remaining to be systematically resolved before we can exploit the full potential of  mixed-precision quantization:

\begin{itemize}
    \item How to define the importance of parameters (i.e. elements in weight and KV-Cache)?
    \item How accurate are required for important parameters and unimportant parameters respectively?
    \item Are important parameters always important?
\end{itemize}

In this paper, we introduce a new criterion, designated as “precision alignment”, to holistically answer the first two questions. The criterion is inspired by the uncertainty calculation commonly used in physics \citep{UncertaintyCalculate}. When performing an addition operation, the uncertainty of the result is close to the bigger one of the uncertainties of two attends, while the smaller one has a much less impact. Therefore, the most economical solution is to align the uncertainties of the two addends to the bigger one. By applying the criterion to the addition operation in matrix multiplication, we can reach a solution to the first two questions: the accuracy of a parameter should be chosen to support a consistent uncertainty between parameters involved in addition operations.

For the third question, in general, our analysis tends to give a negative answer, although for many cases the answer can be yes. A similar conclusion is reached by \cite{dong2024qaq}. In summary, although there are strategies to predict future importance \citep{xiao2023efficient, sheng2023flexgen}, static quantization methods have inherent limitations. In this work, we develop a dynamic KV-Cache quantization scheme to avoid these limitations. 

Based on the precision alignment criterion, we develop a quantization framework that systematically evaluates the importance of parameters and dynamically determines the number of bits to be read. As the first step towards applying the criterion to large model computations, we develop a dynamic KV-Cache quantization scheme that reduces the volume of memory access for the KV-Caches by 25\% and achieves an up to 1.3× speedup in Attention Block with almost no loss of precision during decoding phase of LLM. Our contributions are as follows:

\begin{enumerate}
\item 
We proposed a metric for quantitatively evaluating the importance and precision of each parameter in mixed-precision quantization. To the best knowledge of the authors, this is the first work to enable a quantitative measurement in this area.
\item 
We proposed a dynamic quantization scheme that allows for on-demand data quantization without the need to predict the future importance of data online. The key idea of our work is to read only the bits needed for parameters in KV-Caches during the calculation of LLM’s Attention Block. For some parameters, only the first 11 or 12 bits are necessary to guarantee the accuracy of the result, although they are saved in the Float16 format.
\item 
We found that the columns of K are not equally important, and the rows of V are similarly varied in importance. Our observations suggest that different columns of K should be quantized to different bit widths based on their importance, and rows of V follow the same patterns.
\end{enumerate}

\section{Related Works}

\subsection{LLM Quantization with Outliers}

Model quantization is a technique commonly used to reduce memory usage and accelerate inference speed when deploying large models. During quantization, outliers have a greater impact on the result and require higher precision than normal parameters \citep{dettmers2022gpt3}. To deal with these outliers, some approaches \citep{li2023llm, lee2024owq, dettmers2023spqr, kim2023squeezellm} choose to treat outlier and normal weights in a different manner. They decompose the weight matrix into a dense quantified matrix to store the regular parameters and a sparse high-precision matrix to store the outliers, then calculate with them separately. Besides, some methods blend outliers into normal parameters. Rotation \citep{lin2024rotation} treats matrices as a combination of vectors and randomly rotates them to hide outliers behind normal values. SmoothQuant \citep{xiao2023smoothquant} and AWQ \citep{lin2024awq} scale outliers to fit other elements and multiply them by coefficients after dequantization.

However, the methods above can only identify important parameters through qualitative analysis and manual experiments. Only the largest elements are selected as outliers to retain full precision, lacking quantitative support. On the other hand, methods that analyze the importance of parameters quantitatively \citep{park2024any, kloberdanz2023mixquant} only focus on the importance of a matrix or a layer from a macroscopic perspective. To the best of the authors’ knowledge, this is the first work to systematically analyzes the importance of every parameter in a quantitative manner.

\subsection{KV-Cache Compression and Quantization}

KV-Cache is an essential mechanism to save computations, but at the expense of huge storage consumption and memory access latency. It's necessary to reduce the expense. Early works \citep{liu2024scissorhands, ge2023model, zhang2024h2o} evites useless tokens to save memory. Some methods \citep{zhang2024pyramidkv, yang2024pyramidinfer} based on this way find that the required number of tokens decreases as the layer becomes deeper. Some quantization methods like KIVI \citep{liu2024kivi} also take different measures to quantize KV-Cache to four or even two bits. KIVI suggests that the key cache should be quantified per channel, and the value cache should be quantified per token.

As the self-attention mechanism naturally assigns importance to tokens through the attention score, mixed-precision quantization is introduced for the quantization of KV-Cache. There are some methods \citep{yang2024no, he2024zipcache, yue2024wkvquant, liu2024intactkv} quantize tokens to different precision based on their scores, while other methods \citep{kang2024gear, dong2024qaq} pick outliers out of matrices and save them with high precision. 

Currently, most methods aim to save memory occupied by KV-Cache. The latency associated with quantization and dequantization, nevertheless, is little attended. This paper is the first work targeting to reduce the memory access latency of KV-Cache. As the memory latency is higher than the computational latency by order of magnitude \citep{LLM-KVCache}, our work proves that KV-Cache latency is worth significant optimization effort.

\section{Precision Alignment Criterion}

\subsection{Precision Alignment in Uncertainty Calculation}

In physics, the measurements of physical quantities generally cannot be exact. Therefore, in numerical representation, they are usually expressed in the form of $number\pm uncertainty$, indicating that the result falls within this range. Physics has also established a set of rules for the calculation of such numerical values, which is called uncertainty calculation \citep{UncertaintyCalculate}.

Uncertainty calculation is closely related to LLM quantization. From such a perspective, we can represent the pre-quantified matrix W as the sum of a quantified matrix $W_{quant}$ and an uncertainty $\Delta W$. Therefore, we can express the matrix multiplication of the activation values A and the quantified weights W in the following form of uncertainty computation ($QK^T$ and $SV$ follow the same pattern):
$$AW = A\left( {W_{quant} \pm \mathrm{\Delta}W} \right)$$

We can consider the matrix multiplication above as a combination of two fundamental operations: 

\begin{itemize}
    \item Multiplication of a number from A and a number from W, which corresponds to scalar multiplication in uncertainty calculation:

    $$s \times (x \pm \mathrm{\Delta}x) = sx \pm s\mathrm{\Delta}x$$
    
    \item Addition of the multiplication results, which corresponds to addition in uncertainty calculation:

    $$(x \pm \mathrm{\Delta}x) + (y \pm \mathrm{\Delta}y) = (x + y) \pm (\mathrm{\Delta}x + \mathrm{\Delta}y)$$
    
\end{itemize}

In uncertainty addition, we can find that the uncertainty of the result equals $\Delta x + \Delta y$. On the other hand, the cost of representing $x$ is proportional to $-\log \Delta x + constant$, because we can reduce the uncertainty to $0.5 \Delta x$ when we use one more bit on $x$'s representation. So we can keep the uncertainty of the result in $\Delta x + \Delta y$ at the cost of $\alpha(-\log \Delta x -\log \Delta y + constant)$, which equals $\alpha(-\log (\Delta x\Delta y) + constant)$.

We have the mean value inequality $\Delta x + \Delta y \geq 2\sqrt{\Delta x\Delta y}$, where the conditions for equality is $\Delta x = \Delta y$. So we conclude that the addition is most efficient when $\Delta x = \Delta y$. We call this the “precision alignment criterion”.

\subsection{Precision Alignment in Floating-Point Addition}

Floating-point numbers are widely used in computers. Particularly, in deep learning, the float16 (also known as half-precision floating point, FP16) is the most frequently used format. The float16 format consists of 1 sign bit, 5 exponent bits, and 10 mantissa bits, as shown in Figure 1.

\begin{figure}[h]
  \centering
  \includegraphics[width=\columnwidth]{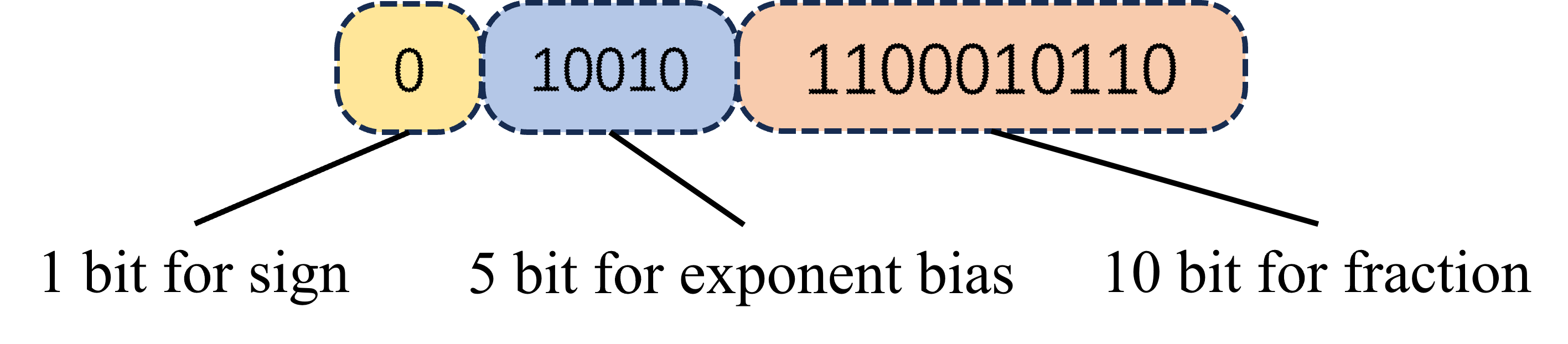}
  \caption{The format of float16}
\end{figure}

The principle outlined in the previous subsection applies to floating point addition if we treat uncertainty as the precision of floating point. This is more comprehensible in the context of vertical addition, as shown in Figure 2. The figure demonstrates a case in which the precisions of two addends are different. We can find that the last few digits of the second addend do not have corresponding positions for the first addend, nor do they have corresponding positions for the results. Thus, these digits are wasted for their absence in result computation.

\begin{figure}[h]
  \centering
  \includegraphics[width=\columnwidth]{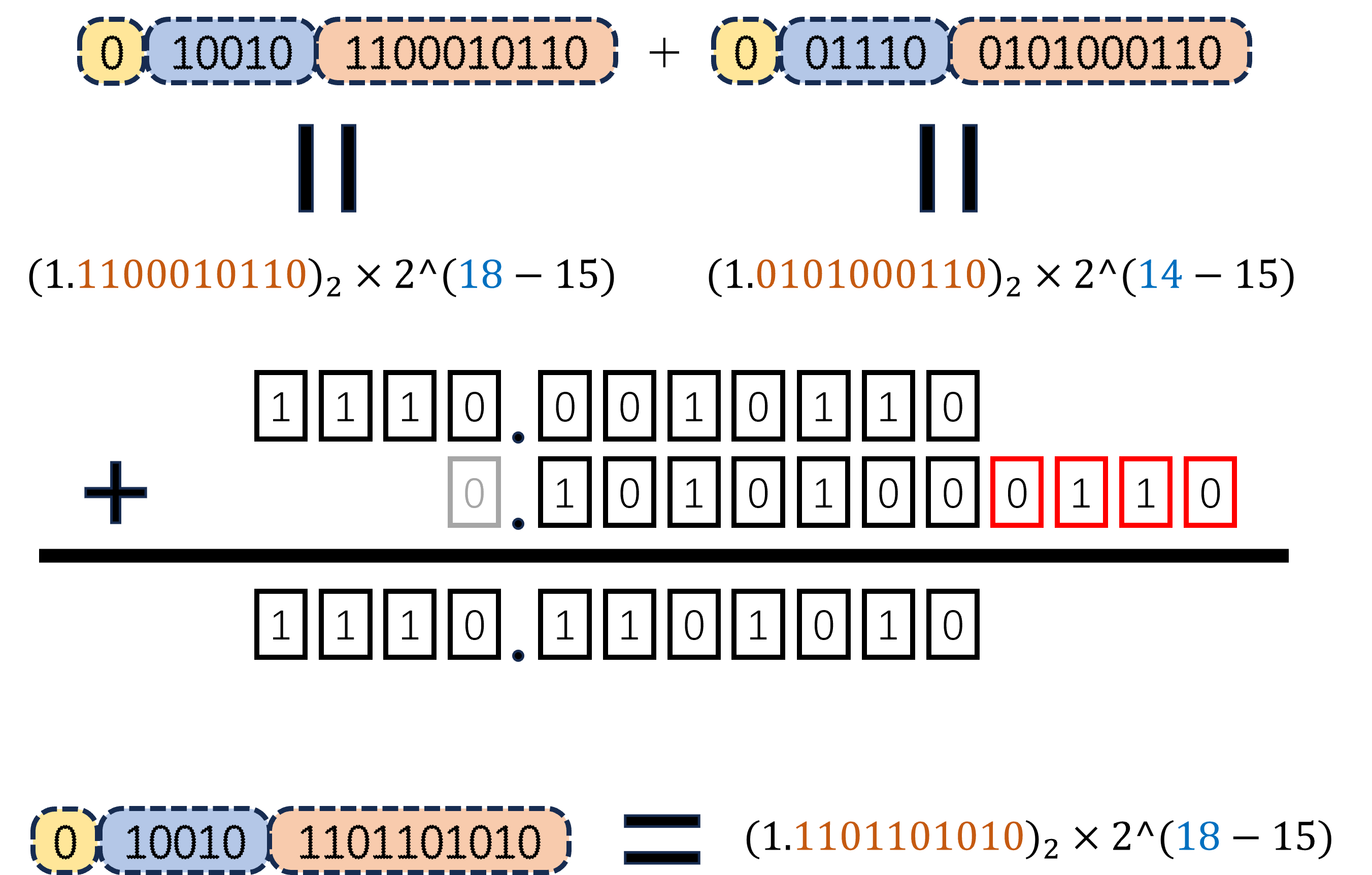}
  \caption{Addition with float16 datatype}
\end{figure}

To avoid loading such wasted bits, we want the following two conditions to be held:

\begin{itemize}
    \item \textbf{Precision Alignment on Attends:} We expect the addends to have the same number of decimal places. Otherwise, some digits have no digits to add. 
    \item \textbf{Precision Alignment on Results:} We expect each addend to have identical numbers of decimal places with the final result. Otherwise, some digits cannot be stored in the result.
\end{itemize}

These two expectations are equivalent in the vast majority of cases. They simply come from different perspectives.

\subsection{Precision Alignment in GEMM}

As mentioned earlier, matrix multiplication consists of two fundamental operators---scalar multiplication and addition. We can understand matrix multiplication from a 3D perspective, as Basil Hosmer described \citep{3DGEMM}. Matrix multiplication corresponds to a large cube, with the two matrices involved in the multiplication and the result corresponding to the three faces of the cube, as Figure 3 shows. Based on this correspondence, the scalar multiplication operation represents the smaller cubes, while summing up the results of the scalar multiplications corresponds to the addition between the smaller cubes in the horizontal dimension.

\begin{figure}[h]
  \centering
  \includegraphics[width=\columnwidth]{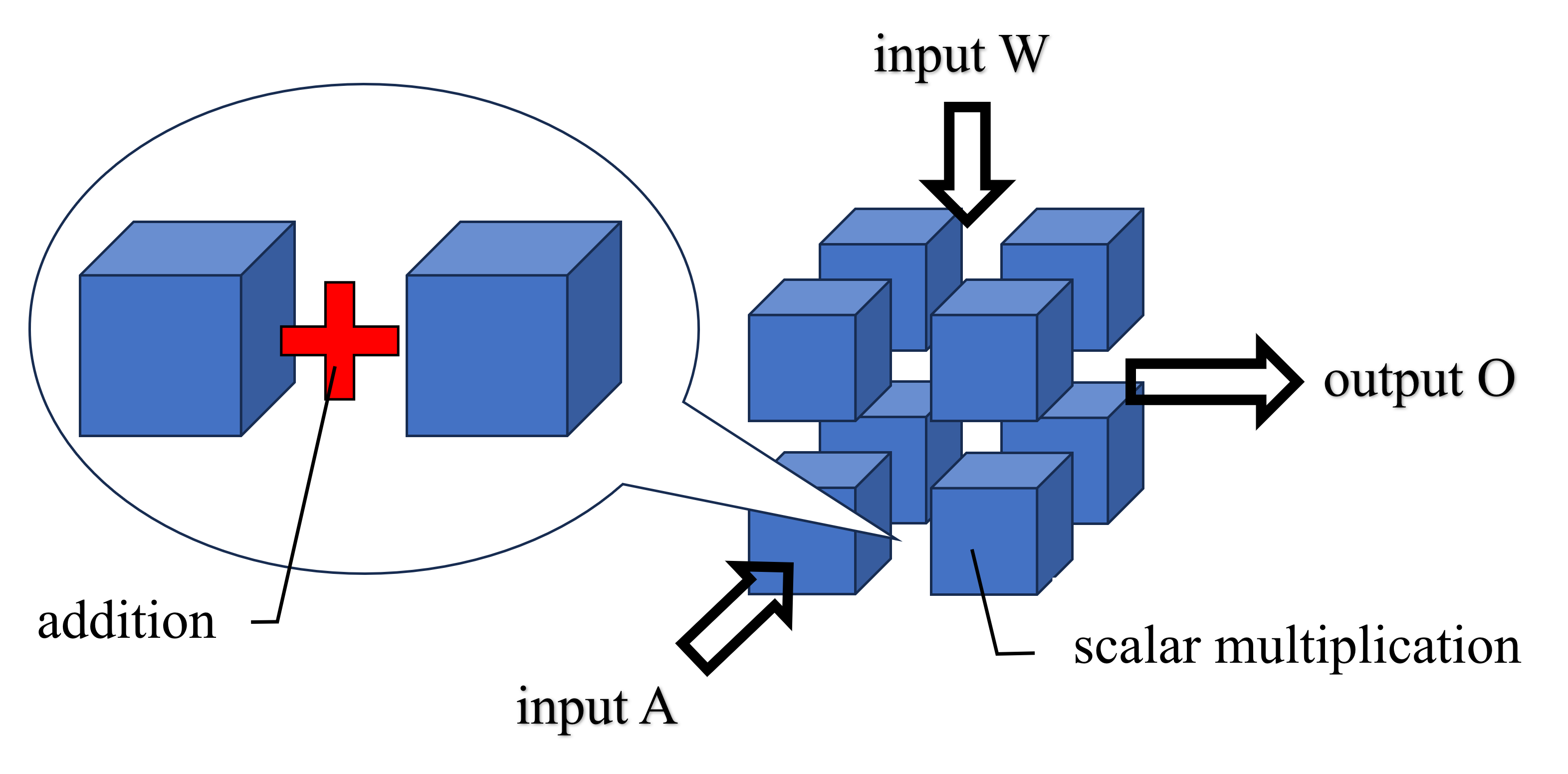}
  \caption{GEMM from 3D prospective}
\end{figure}

The previous subsection suggests that addition with consistent precision among the addends is the most efficient. This principle is also suitable in additions inside General Matrix Multiply (GEMM). We can treat the addition in GEMM from two perspectives:

\begin{itemize}
    \item \textbf{Precision Alignment on Attends:} The precision of the intermediate results in GEMM should be consistent.
    \item \textbf{Precision Alignment on Results:} The precision of each intermediate result in GEMM should be the same as that of the final result.
\end{itemize}

\section{A Framework for Aligning Precision}

In the previous text, we proved that it’s preferable to let addends align with each other and the final results. Based on this criterion, we develop a framework to propose the precision of each parameter in weight and KV-Cache so that all the additions are aligned with precision.

To apply the principle of precision alignment, we must be aware of certain information about the intermediate results and the final outcomes. It should be noted that we only need the approximate magnitudes of these values, which can be estimated before the actual computation. In the following discussion, we will assume that we have already obtained the magnitudes of these values.

\begin{figure}[h]
  \centering
  \includegraphics[width=\columnwidth]{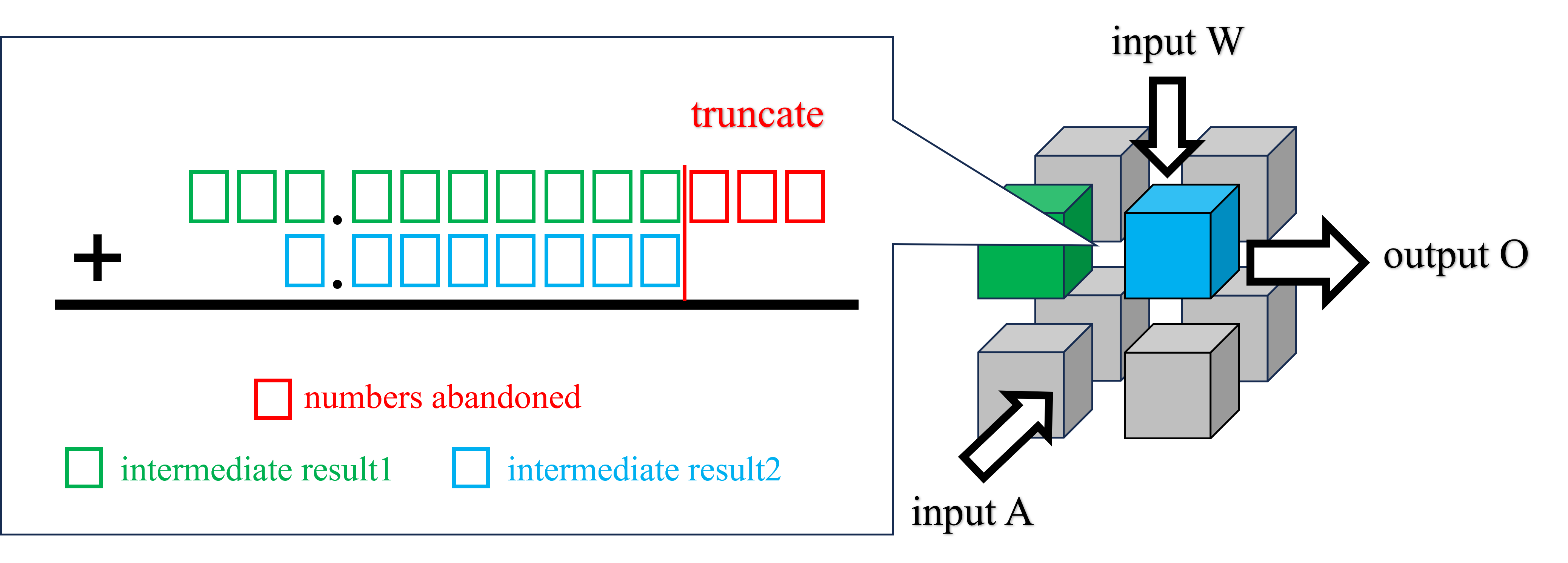}
  \caption{Attend Precision Alignment}
\end{figure}

\textbf{Rule 1 (Attend Precision Alignment):} We can consider the problem from the perspective of addend alignment. The objective is to keep consistent precision across all intermediate results. This means when there is inconsistency in precision, we can slightly reduce the precision of some higher-precision elements to achieve uniformity in the precision of the intermediate results. The final effect of this adjustment is that the precision of all intermediate results aligns with the lowest precision among them, as illustrated in Figure 4.

\begin{figure}[h]
  \centering
  \includegraphics[width=\columnwidth]{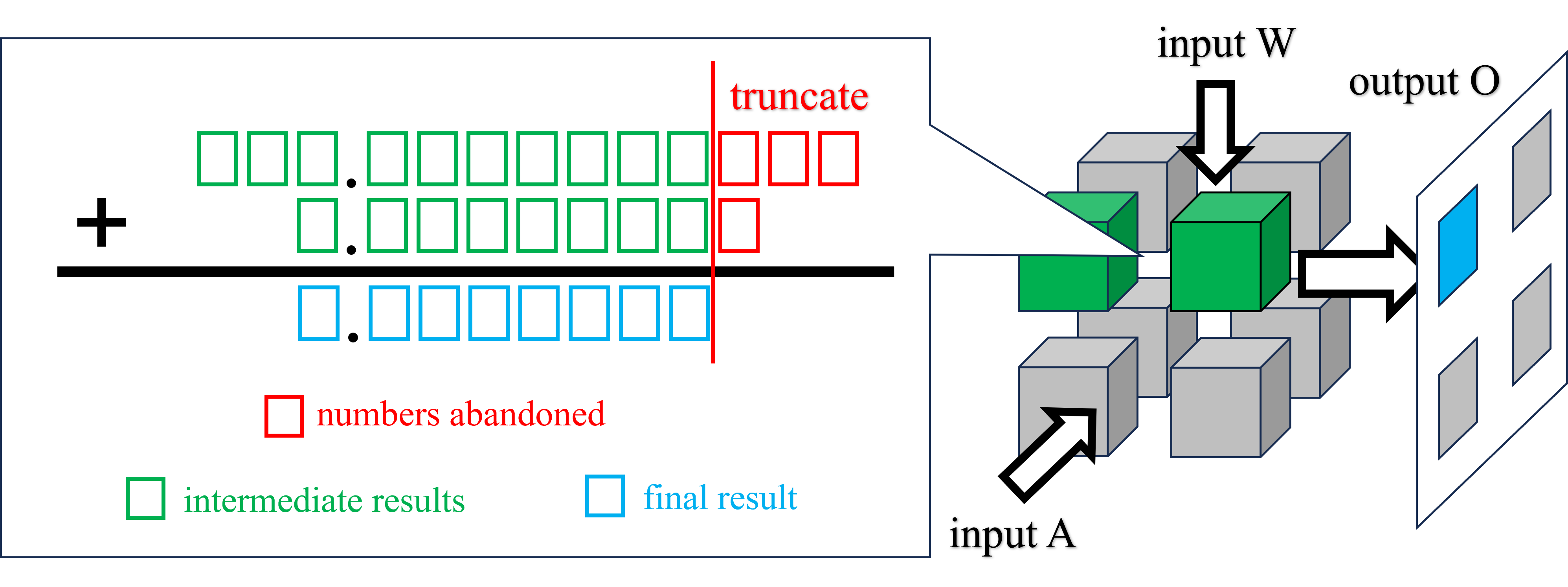}
  \caption{Result Precision Alignment}
\end{figure}

\textbf{Rule 2 (Result Precision Alignment):} We can also consider this problem from the perspective of result alignment. Here we aim for the precision of all intermediate results to be same as the precision of the final result. If any intermediate result has a precision higher than the final result, we can reduce it to match the final result's precision, as shown in Figure 5.

Both perspectives offer a solution to determine the precision of the intermediate results. As mentioned earlier, these two forms are equivalent, allowing us to choose one according to the computation scenario. When the results can be estimated more accurately, we can use Rule 2 to align the precision of intermediate results. Otherwise, when the results are difficult to estimate and the number of addends is relatively small and manageable, we can choose Rule 1.

Once we obtain the precision of the intermediate results, we can infer the precision of the weights (K-Cache, V-Cache). Since scalar multiplication does not change the relative uncertainty (uncertainty/value), and the quantization bit width is only related to the relative uncertainty of the data, we can directly use the bit width of the intermediate results as the quantization bit width for the weights (K-Cache, V-Cache).

So far we have proposed a framework to determine the bit width of quantization. In the next section, we will introduce an application on KV-Cache using our framework, named “AlignedKV”. This work devotes to cutting down the total memory access of KV-Cache and reducing the memory latency of this part, which is the bottleneck of LLM inference speed.

\section{AlignedKV: Quantizing KV-Cache with Precision Alignment Criterion}

\subsection{Dynamic Quantization}

Currently, the vast majority of quantization methods are static, with only a few works employing dynamic quantization based on data \citep{wang2021spatten} as most quantization techniques aim to save storage space, and hence a static quantization suffices.

However, in the context of KV-cache, dynamic quantization offers compelling advantages. Firstly, dynamic quantization can reduce memory access latency without adding additional delays. It's important because memory access latency is the bottleneck for the inference speed in this stage. Secondly, static quantization introduces a significant problem for the usage of mixed-precision quantization in KV-cache, as it requires determining the importance of elements at the time of quantization. Although there are numerous methods available to predict the importance of elements to address this issue (i.e., the persistence of importance hypothesis)\citep{ge2023model, liu2024scissorhands}, these predictions are always subject to counterexamples \citep{dong2024qaq}.

We propose a method to dynamically quantize KV-cache, which quantizes data when it is used. When storing the KV-cache, we maintain additional data structures to store the approximate magnitudes of the KV-cache elements, which is one of the preconditions for precision alignment. Afterwards, during the computation of $QK^T$ and $SV$, we use the precision alignment criterion to determine the required precision for each element in K and V. 

\begin{figure}[h]
  \centering
  \includegraphics[width=\columnwidth]{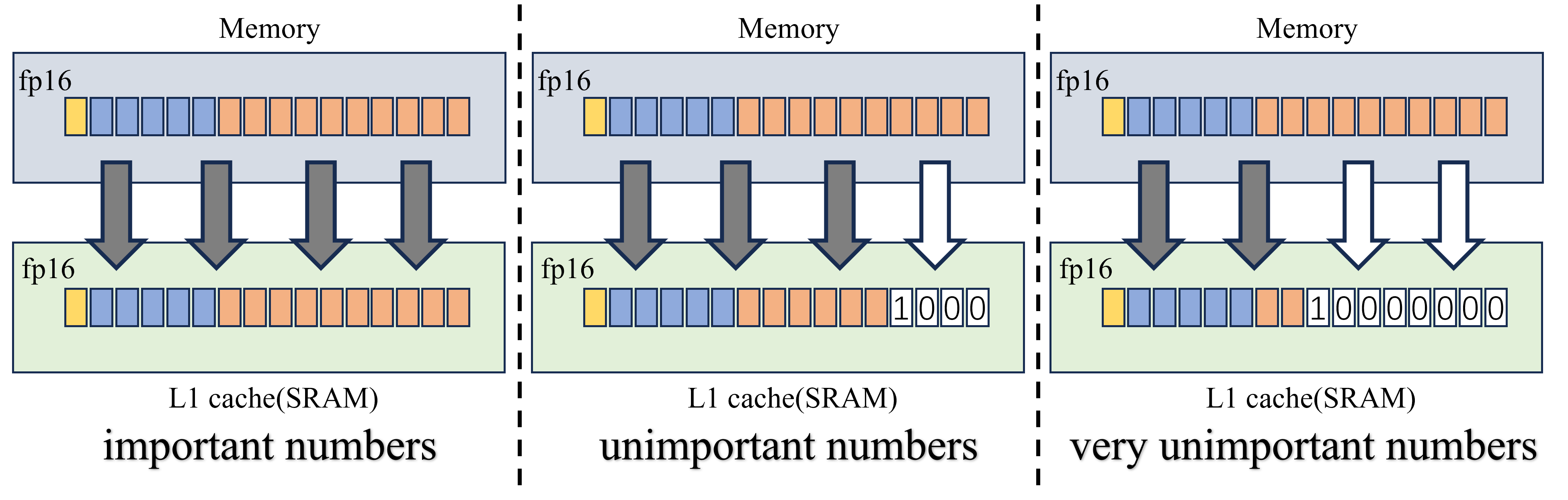}
  \caption{Dynamically truncate parameters with different importance while loading them from memory}
\end{figure}

\begin{figure*}[t]
  \centering
  \includegraphics[width=\textwidth]{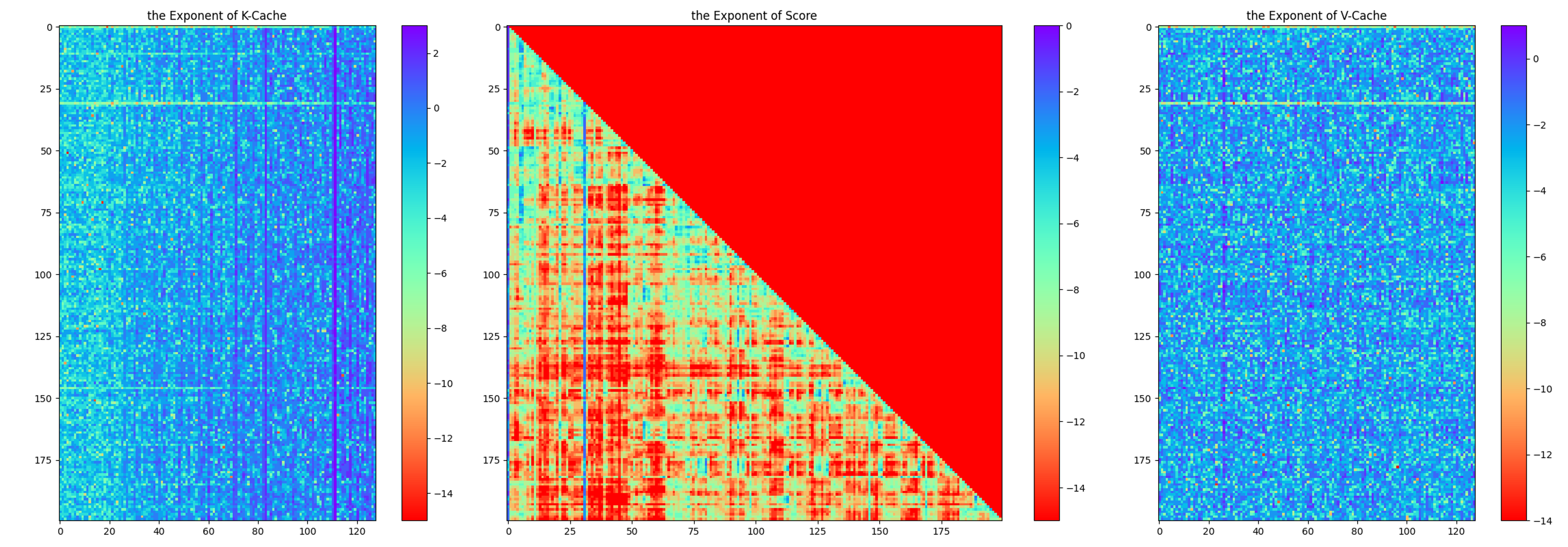}
  \caption{The distribution of exponent bits for K-Cache (left), Score (middle) and V-Cache (right)}
\end{figure*}

We then retrieve only the necessary number of bits from memory on an as-needed basis, as illustrated in Figure 6. According to the results of the precision alignment principle, we read all 16 bits for important parameters, the first 12 bits for less important parameters, and the first 8 bits for particularly unimportant parameters. The remaining bits are filled with 100... to minimize the precision loss to the greatest possible extent.

\begin{figure}[h]
  \centering
  \includegraphics[width=\columnwidth]{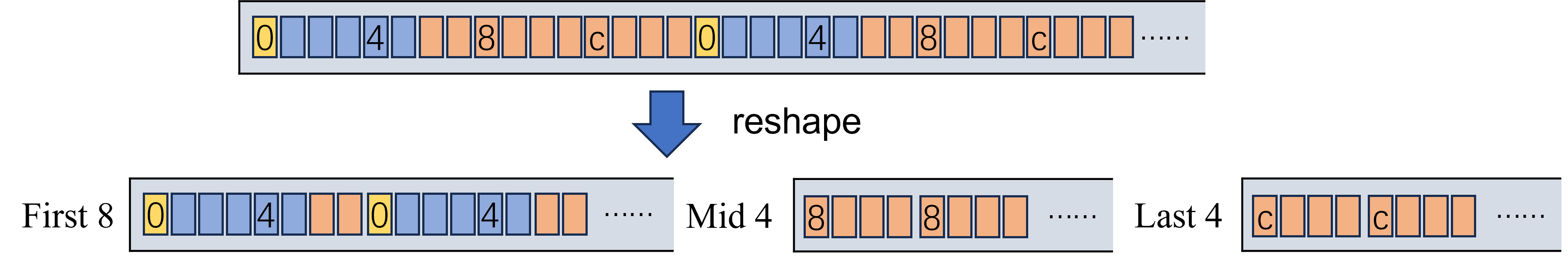}
  \caption{Data rearrangement for hardware friendliness}
\end{figure}

At the same time, as the GPU hardware can only read memory at a minimum unit of several bytes, we rearrange the storage structure to ensure continuity for hardware access, as shown in Figure 8.

\subsection{Quantization of K-Cache}

AlignedKV follows the process outlined below, applicable to both K-Cache and V-Cache: 

\begin{enumerate}
\item Obtain the approximate magnitude information of the values in KV-Cache by an additional data structure, preparing the data for precision alignment. We only need the order of magnitude information.
\item Apply our framework to determine the required precision for loading each data element.
\item Perform truncation on the data elements using dynamic quantization methods while loading data from GPU memory to L1 cache.
\end{enumerate}

Among these, the latter two have already been addressed in the previous sections, leaving only the first step, which requires specific strategies based on the characteristics of the data.

Based on observations of the K-Cache, we found that the data distribution exhibits a columnar characteristic, where the data within a single column is relatively similar and the data between different columns varies significantly, as shown in Figure 7. In other words, we can use a single value from a column (specifically, the maximum value) to represent the order of magnitude for that column.

Thus, we maintain the information of the maximum value for each column of the K-Cache, which is recorded as $K_{colmax}$. When computing $QK^T$, we first use the value of $Q$ along with the values of $K_{colmax}$ to estimate the order of magnitude of the intermediate results. Then, we can determine the aligned precision of the intermediate results by Rule 1 (Addend Precision Alignment), which allows us to infer the required bit width for each element in the K-Cache. All of this is derived quantitatively, rather than merely through qualitative analysis.

Then, we load the required bits of each parameter from memory using the proposed dynamic quantization methods. As a result, we can avoid loading a large number of redundant bits from GPU memory, which significantly reduces the memory access latency for this part.

\subsection{Quantization of V-Cache}
\begin{table*}[t]
    \centering
    \small
    \renewcommand\arraystretch{1.6}
    \begin{tabular}{|c|c|c|c|c|c|c|c|}
    \hline
        \multirow{2}{*}{Type} & \multirow{2}{*}{Method} & \multicolumn{6}{c|}{Proportions of each Relative Error Range} \\ \cline{3-8}
        & & 0 & (0,1/1024) & [1/1024, 1/512) & [1/512, 1/256) & [1/256, 1/128) & [1/128, inf) \\ \hline
        \multirow{2}{*}{$QK^T$} & AlignedKV & 56.30\% & 37.00\% & 5.42\% & 0.73\% & 0.31\% & 0.25\% \\ \cline{2-8}
        & Truncated to 13 bits & 18.65\% & 32.70\% & 36.21\% & 10.26\% & 1.36\% & 0.83\% \\ \hline
        \multirow{2}{*}{$SV$} & AlignedKV & 76.12\% & 18.14\% & 3.61\% & 1.29\% & 0.39\% & 0.44\% \\ \cline{2-8}
        & Truncated to 13 bits & 20.04\% & 29.47\% & 26.66\% & 13.82\% & 5.71\% & 4.30\% \\ \hline
    \end{tabular}
    \caption{The distribution of relative error for result of AlignedKV (compare with normal result without any quantization)}
\end{table*}

Similar to the quantization of K-Cache, the quantization of V-Cache also requires obtaining approximate order of magnitude information for the addends or results in order to apply precision alignment criterion. However, unlike K-Cache, V-Cache does not exhibit a columnar distribution characteristic, which prevents us from applying the quantization method used for K-Cache. Fortunately, we observed that the magnitude differences among the elements in Score matrix (recorded as S) are quite significant, as shown in Figure 7, which means that we only need to compute the product of a subset of the larger elements in S with the corresponding V to obtain an approximate estimation of final results.

We first select the top k largest values from the score matrix S using an approximate top-k algorithm (we don’t use the exact top-k algorithm for speed) and multiply them by their corresponding V values to obtain an estimate of the SV result, referred to as $O_{est}$. Subsequently, we can determine the precision alignment of the intermediate results by Precision Alignment on  Result using $O_{est}$. Similar to the quantization of K-Cache, the alignment of the intermediate results allows us to infer the required bit width for each element in V-Cache. We can then reduce memory access latency by reading only the necessary bits for each element.

\section{Experiments}

\subsection{Experiments Setup}

We performed comprehensive experiments to validate the effectiveness of the proposed method. We conducted our experiments on the Llama-2-7b model with our implementation of KV-cache coded in CUDA. The experiments were conducted on a Nvidia V100 GPU. It must be noted that the proposed method is friendly to efficient hardware implementation and the dedicated hardware can be significantly more efficient. 

\subsection{Reduction in Bit Widths}

\begin{figure}[h]
  \centering
  \includegraphics[width=\columnwidth]{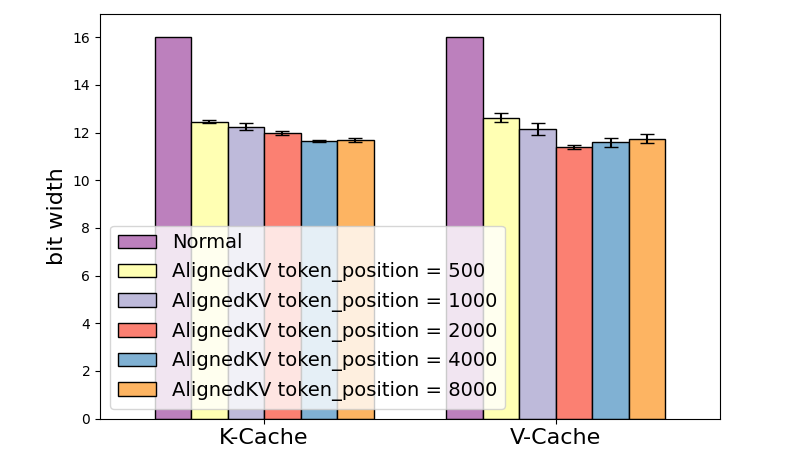}
  \caption{Average bit width of KV-Cache}
\end{figure}

In this experiment, we apply AlignedKV on the actual KV-Cache data generated by Llama-2-7B. We statistically analyze the alignment of actual accuracy and the number of bits required for each element in the KV-Cache. The results are illustrated in Figure 9. They clearly demonstrate that after using AlignedKV, the average bit width for each element decreases from 16 to approximately 12, resulting in a significant reduction of the memory access latency during the KV-Cache access. It drops with the increase of context length, indicating that our method has significant potential for longer context.

\subsection{Accuracy of GEMM Result}

In quantization, it is essential to consider not only the compression rate but also the accuracy. In the previous experiment, we also analyze the impact of applying AlignedKV on the accuracy of the results. We record the changes in $QK^T$ and $SV$ GEMM after using AlignedKV and represent these changes through the distribution of relative errors.

We use the following formula to calculate the relative error for each value in the results:

$$
{\rm relative\ error}=\frac{|R_{AlignedKV}-R_{normal}|}{R_{normal}}
$$

\begin{align*}
    {\rm where} \  & R_{AlignedKV} = QK_{AlignedKV}^T {\rm \ or\ } SV_{AlignedKV} \\
    {\rm and} \  & R_{normal} = QK_{normal}^T {\rm \ or\ } SV_{normal}
\end{align*}

We calculate the relative error for each value in the result matrix and observe their distribution, as shown in Table 1. Notably, although AlignedKV reduces the bit widths of KV-Cache by 4 bits, most of the results remain unchanged. In contrast, uniformly removing 3 bits from each element in KV-Cache yields significantly poorer results, despite this approach achieving a lower compression rate than AlignedKV. This indicates that by quantitatively analyzing the bit-width requirements for each element and letting them achieve a state of “alignment”, we can take both a high compression rate and high accuracy.

\subsection{Runtime of AlignedKV}

We also evaluated the actual runtime. We run the inference of a Llama-2-7B model with 128-token input and up to 8192-token completion (disregarding the model’s maximum context length of 4096 because we only care about the time cost). To highlight the effectiveness of AlignedKV, we only focus on the computations involved in the KV-Cache component. The computations are as follows:

\begin{align*}
    S&=QK^T/\sqrt{128} \\
    S&=softmax(S) \\
    O&=SV
\end{align*}

When the context length reaches hundreds of thousands or even millions, the memory access latency of the KV-Cache will become the bottleneck for the entire model's running speed. At this point, the benefits of AlignedKV will be substantial.

We compare AlignedKV with the native implementation in Torch and two other quantization methods---KIVI \citep{liu2024kivi} and GEAR \citep{kang2024gear} (these quantization methods are set to 10 bits for fairness). We also set a control group using AlignedKV's code to evaluate the memory access reduction AlignedKV brings, ignoring the additional overhead associated with CUDA program execution. The only difference between the control group and AlignedKV is that the latter always loads elements as 16-bit values.

\begin{figure}[h]
  \centering
  \includegraphics[width=\columnwidth]{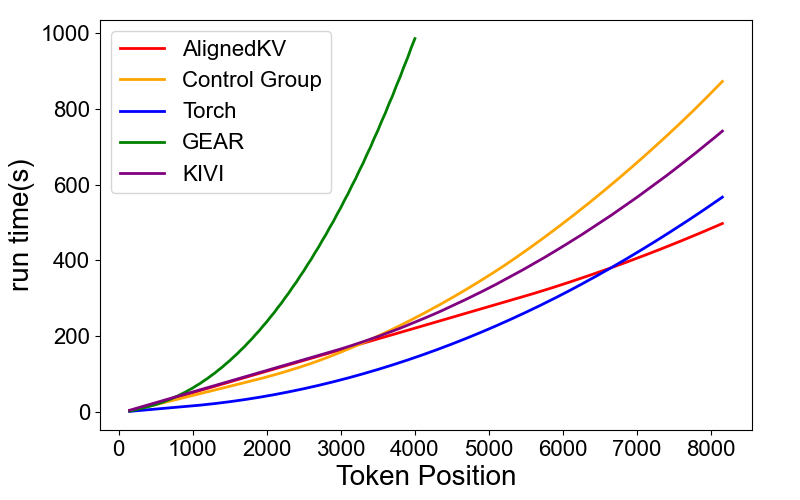}
  \caption{Run time of AlignedKV and other methods}
\end{figure}

The runtime results are displayed in Figure 10. With the increasing position of tokens, the length of the KV-Cache also grows, at which point the advantages of AlignedKV in reducing memory access latency become apparent. When the generating length reaches around 7000 to 8000, AlignedKV demonstrated the shortest execution time among all experimental groups, surpassing the performance of the native implementation in Torch. The CUDA implementation of Torch is highly optimized. We expect a higher level of performance improvement of our implementation after optimization.

\subsection{End-to-End Accuracy Performance}

We are willing to present the end-to-end accuracy performance of AlignedKV, as AlignedKV doesn't cause any loss of precision. We benchmark it on CoQA\citep{reddy2019coqa}, TruthfulQA\citep{lin2021truthfulqa}, and GSM8K\citep{cobbe2021gsm8k} tasks using default settings, as KIVI does. We utilize an open-source evaluation repository \citep{eval-harness} to conduct this experiment. The results are shown in Table 2.

\begin{table}[h]
    \centering
    \small
    \renewcommand\arraystretch{1.6}
    \begin{tabular}{|c|c|c|c|c|}
    \hline
        Model & Method & CoQA & TruthfulQA & GSM8K \\ \hline
        \multirow{3}{*}{Llama-2-7B} & origin & 63.88 & 32.31 & 13.80 \\ \cline{2-5}
        & AlignedKV & 63.88 & 32.31 & 13.95 \\ \cline{2-5}
        & KIVI & 64.42 & 33.90 & 12.66 \\ \hline
    \end{tabular}
    \caption{Performance comparison between AlignedKV and the origin model}
\end{table}

Through experiments, we find the results of AlignedKV are very close to the origin model, which means AlignedKV has no impact on the result of LLMs. The results are as we expected since the errors in GEMM results will be diluted on a higher level, and the results of GEMM by AlignedKV are already accurate.

\begin{table}[h]
    \centering
    \footnotesize
    \renewcommand\arraystretch{1.6}
    \begin{tabular}{|c|l|}
    \hline
        Method & \multicolumn{1}{|c|}{Output} \\ \hline
        origin & who is at the front. I am very glad to hear that he... \\ \hline
        AlignedKV & who is at the front. I am very glad to hear that he... \\ \hline
        KIVI & who is \textcolor{red}{in the army, and who is now in the hospital...} \\ \hline
    \end{tabular}
    \caption{Model output with the prompt “I have just received a letter from my brother,”}
\end{table}

To further demonstrate that our method incurs exactly no loss in end-to-end accuracy, we used the model to generate sentences by greedy search. We found that the results generated by AlignedKV are completely consistent with those produced by the original model. In contrast, although the sentences generated by KIVI are coherent, the initial words differ from those generated by the original model. A brief example is shown in Table 3.

\section{Conclusion}

Currently, a quantitative theoretical framework is greatly needed for the mixed-precision quantization methods, which still rely on experiments to select a preferable bit-width. To address this issue, we proposed a Precision Alignment criterion, which makes it feasible to derive the optimal bit-width for each parameter through theoretical analysis rather than relying solely on experimentation. This approach allows us to achieve both high compression rates and high accuracy. 

Furthermore, based on this theory, we implemented a dynamic quantization method, which quantizes the KV-Cache to reduce memory access latency. This method can significantly accelerate the inference speed of the Attention component during the decoding phase with no impact on the accuracy of the final result.

\section{Acknowledgments}

This work is supported by Sunlune Company. We sincerely thank all the reviewers for their valuable feedback and suggestions.

\bibliography{aaai25}

\end{document}